\documentclass[letterpaper]{article} 
\usepackage{aaai23}  
\usepackage{times}  
\usepackage{helvet}  
\usepackage{courier}  
\usepackage[hyphens]{url}  
\usepackage{graphicx} 
\usepackage{natbib}  
\usepackage{caption} 
\usepackage{algorithm}
\usepackage{algorithmic}

\usepackage{newfloat}
\usepackage{listings}
\usepackage{xcolor,colortbl}
\usepackage{kotex}
\usepackage{amsmath}
\usepackage{booktabs}
\DeclareCaptionStyle{ruled}{labelfont=normalfont,labelsep=colon,strut=off} 
\floatstyle{ruled}
\newfloat{listing}{tb}{lst}{}
\floatname{listing}{Listing}
\title{Leveraging Skill-to-Skill Supervision for Knowledge Tracing}
\author{
    Hyeondey Kim \textsuperscript{\rm 1*}, Jinwoo Nam\textsuperscript{\rm 1*}, Minjae Lee\textsuperscript{\rm 1*}, Yun Jegal\textsuperscript{\rm 1}, Kyungwoo Song\textsuperscript{\rm 2}\\
}
\affiliations{
    \textsuperscript{} Turing AI Research
    \\
    \textsuperscript{\rm 1} Turing, 431, Teheran-ro, Gangnam-gu, Seoul, Republic of Korea  \\
    \textsuperscript{\rm 2} University of Seoul, 163 Seoulsiripdaero, Dongdaemun-gu, Seoul, Republic of Korea  \\
    \{sieg, matthew, marcus, tony\}@teamturing.com, kyungwoo.song@uos.ac.kr
}

\usepackage{bibentry}

\definecolor{blue}{rgb}{0, 0, 0.9}
\definecolor{red}{rgb}{1, 0, 0}
\definecolor{green}{rgb}{0.0, 0.7, 0.0}

\begin{document}

\maketitle

\begin{abstract}

Knowledge tracing plays a pivotal role in intelligent tutoring systems. This task aims to predict the probability of students answering correctly to specific questions. To do so, knowledge tracing systems should trace the knowledge state of the students by utilizing their problem-solving history and knowledge about the problems. Recent advances in knowledge tracing models have enabled better exploitation of problem solving history. However, knowledge about problems has not been studied, as well compared to students' answering histories. Knowledge tracing algorithms that incorporate knowledge directly are important to settings with limited data or cold starts. Therefore, we consider the problem of utilizing skill-to-skill relation to knowledge tracing. In this work, we introduce expert labeled skill-to-skill relationships. Moreover, we also provide novel methods to construct a knowledge-tracing model to leverage human experts' insight regarding relationships between skills. The results of an extensive experimental analysis show that our method outperformed a baseline Transformer model. Furthermore, we found that the extent of our model's superiority was greater in situations with limited data, which allows a smooth cold start of our model.
\end{abstract}

\section{Introduction}

\begin{figure*}[t!] 
    \centerline{\includegraphics[width=1.0\textwidth]{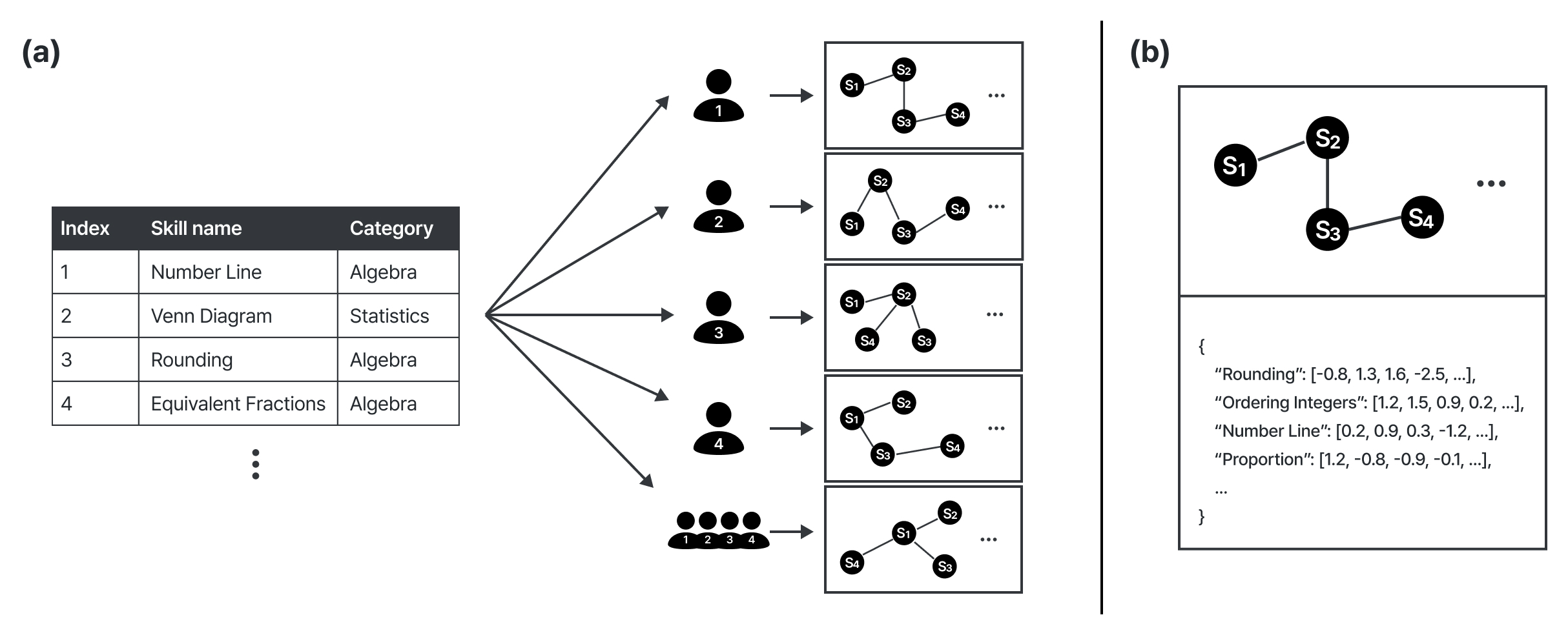}}
    \caption{(a) Overall process of labeling skill-to-skill relationships. (b) result of converting graph to embeddings.}
	\label{fig:dataset}
\end{figure*}

Due to the recent quarantine, online learning has been a major issue for student achievements. As knowledge tracing is a crucial component for better online classes, it has attracted more attention in recent years. Knowledge tracing is a task that aims to predict the future performance of a student by discovering the state of their knowledge. To do so, knowledge tracing models should be able to discover meaningful information from problem solving history and problem metadata.

Recent advances in deep knowledge tracing are noteworthy. 
Starting from a simple sequential model \cite{dkt}, many deep knowledge tracing methods have been proposed, inspired by graph neural networks, item-response theory, and attention mechanisms\cite{seq2seqattn}. In particular, contemporary works inspired by the Transformer model\cite{transformer} have showing significantly better prediction performance compared to previous methods.

However, these advances mostly address utilizing information on problem-solving history, not knowledge itself.
In the real world, there are plenty of other useful information or features other than problem-solving history. For example, information about the problem itself, such as annotated problem difficulty and problem types (multiple-choice or subjective), could be utilized to incorporate prior knowledge. Also, knowledge about how problems are connected is clearly important.

To this end, skill-to-skill relationships are among the most important but relatively rarely studied information. It describes how skills relating to one problem are connected with the skills required to solve another problem. This kind of information is particularly useful in mathematics, a field where skills are very closely linked to one another. We can expect that if a knowledge tracing model can utilize information on skill-to-skill relationships, then it can use students' levels of achievement in specific skills to infer students' expected levels of achievement in strongly related skills. Despite its significance, determining how to use skill-to-skill relationships while constructing a model is unintuitive. Skill-to-skill relationships are represented in the form of a graph, which cannot be easily inserted into existing knowledge tracing models. Moreover, the fact that datasets which support skill-to-skill relationships are uncommon also makes exploiting that knowledge challenging.

Therefore, to facilitate utilizing skill-to-skill information, we propose an expert-labeled skill-to-skill dataset, ASSIST2009-SSR (ASSIST2009 Skill to Skill Relationship), which is a zero-one type description explaining which combinations of skills are strongly related. The ASSIST2009-SSR dataset consists of five different relationship data. Four mathematics experts designated strongly related skill pairs according to each person's opinion. Subsequently, they discussed together until they reached an agreement and selected related skill pairs using their common opinion.

\footnote{{\rm *} These authors contributed to this paper equally.}

In addition, we propose a novel knowledge tracing method that can incorporate expert labeled skill-to-skill relation data. We first used the Node2Vec method to convert the graph-structured skill-to-skill relationships data into skill embedding data. Then we applied projection-loss architecture to add additional skill-inherent information into the embedding data. The empirical results show that exploiting skill-to-skill relations improved the quantitative performance of the knowledge tracing model, particularly when using a small dataset for training.

The contributions of this study are summarized as follows.
\begin{itemize}
    \item To the best of our knowledge, we provide the first dataset with labels created by human experts for skill-to-skill relationships on the ASSIST09 dataset \footnote{https://github.com/weareteamturing/assist09-skill-to-skill-relationships}
    
    \item We propose a novel knowledge tracing method that can exploit skill-to-skill relations labeled by experts. Our method improved the performance of knowledge tracing methods significantly. In particular, the proposed approach showed an AUC improved by 1.35\%p with a dataset of limited size.

    \item We conducted a thorough analysis of how expert-labeled skill-to-skill relation affects knowledge tracing performance. Empirical results show the performance of our method exceeded that of an existing baseline model, especially in a data-scarce situation.
\end{itemize}

\section{Related work}

In this section, we introduce the DKT and DKT with knowledge graph algorithms.
First, methods using algorithms based on deep knowledge tracing \cite{bkt} were widely used to solve knowledge tracing before the development of advanced deep learning models. RNN and LSTM \cite{lstm}-based deep learning models \cite{dkt} have made huge breakthroughs in knowledge tracing. The method incorporates the sequential structure of the problem solving algorithm with the recurrent operation. It may be noted that this approach does not incorporate human knowledge directly.

Second, DKT with knowledge graphs and graph neural network processing for graph-structured data is one of the most common areas of research on knowledge discovery systems \cite{node2vec}. Along these lines, graph neural networks \cite{gnn} were proposed to uncover the relationship of a graph. One of the key purposes of a knowledge tracing model is to identify relationships between skills. Thus, regarding skill-to-skill relationships as a graph is a widely considered approach. Graph neural network-based knowledge tracing models \cite{gkt}, and pre-training the knowledge tracing with a graph representation \cite{pebg} have also been proposed.

Our work involves three improvements compared to the previous works. First, we utilize a Transformer layer to incorporate the sequential history of users' problem solving. Therefore, our method handles the long-term history effectively. Second, we construct the skill-to-skill relationship annotation dataset from diverse experts. The knowledge graph constructed based on the experts' knowledge might attract diverse audiences. Third, we propose a simple yet effective method to incorporate human prior knowledge easily. We apply the Node2vec method to compute the embedding of the skills from the skill-to-skill relationship dataset. Next, we propose a new projection layer-based optimization method to compute the problem embedding that incorporates the knowledge itself as well as the knowledge tracing simultaneously.

\section{Dataset}

\subsection{ASSIST09 dataset}

\begin{figure*}[t!] 
    \centerline{\includegraphics[width=1.0\textwidth]{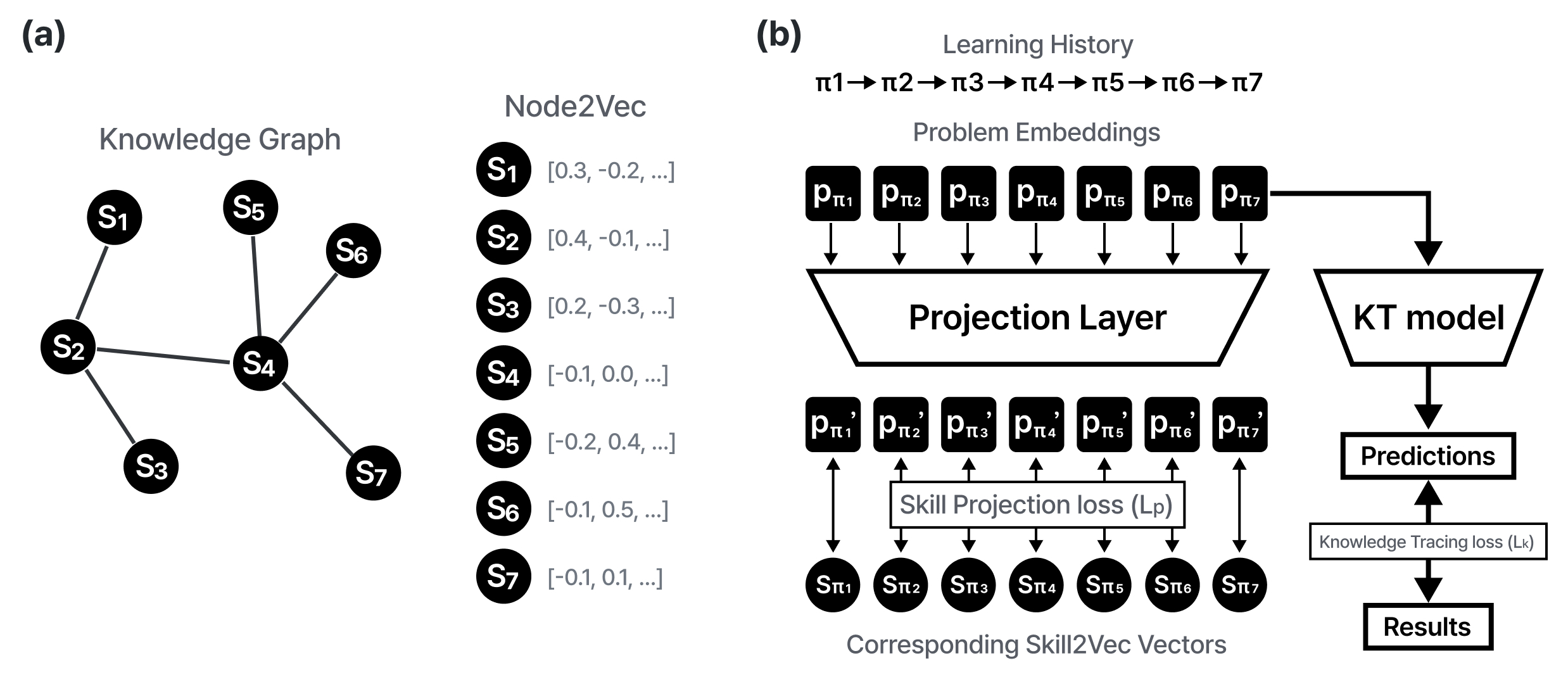}}
    \caption{Proposed knowledge tracing model. (a) Skill2Vec feature generation. (b) Skill Projection loss.}
	\label{fig:overview}
\end{figure*}

The ASSIST09 skill builder dataset is one of the most widely used datasets on knowledge tracing, which is a representative education benchmark dataset gathered from ASSISTments, in which problem sets to are assigned to students to develop their abilities.\footnote{https://sites.google.com/site/assistmentsdata/home/2009-2010-assistment-data/skill-builder-data-2009-2010?authuser=0} 
It provides 525,535 records, and the question tagged one or more skills out of a total of 110 skills in the K-12 mathematics curriculum. Table \ref{tab1} demonstrates the details of the ASSIST09. Although it contains many records, due to the sparsity of the dataset, most of the interactions are concentrated on a few questions. 
Because constructing dense representations for each question is laborious, we considered each problem as a skill in the experiments and focused on how to build informative skill representations. We used 90\%, 10\% of records as the training set and evaluation set, respectively.

\begin{table}[hb]
\caption{ASSIST09 dataset}\label{tab1}
\centering
\begin{tabular}{ll}
\toprule
Attribute &  Numbers \\
\midrule
Number of students & 4,218\\
Number of questions & 26,689 \\
Number of skills & 110 \\
Number of records& 525,535 \\
\bottomrule
\end{tabular}
\end{table}

\subsection{Skill-to-skill relationships}
Four math experts labeled the relationship of skills for ASSIST09. To ensure the quality of the labeling, math experts with bachelor's degrees in mathematics or mathematics education who had worked for at least five years in the education industry were selected. 
ASSIST09 is composed of a total of 110 skills, and math experts labeled each skill pair which they considered to be related. To label the skill-to-skill relationship more accurately, experts selected one of the categories of algebra, probability, statistics, and geometry for 110 skills. 
In addition, each of the four experts independently labeled the relationship of skills to reduce the bias against potential individual biases. 
Subsequently, they labeled skill-to-skill relationships together once more until converging into individual opinions, and we used the converged skill-to-skill relationship opinion dataset in the training process for the model. Figure 1-(a) illustrates a brief overview of the expert labeling process.

\section{Methodology}

Here, we describe the proposed deep knowledge tracing model designed to effectively incorporate skill-to-skill relationships in knowledge tracing. Figure \ref{fig:overview} shows a brief overview of the model.

\subsection{Knowledge Tracing}
The input for knowledge tracing is problem solving history $H=\{(\pi_i,r_i)\}_{i=1}^{N_{seq}}$. Where, $\pi_i$ is the problem that user tried to solve, $r_i \in \{0,1\}$ is whether the user provided correct answer, and $N_{seq}$ is the length of the problem solving history. 

Additionally, we used a problem embedding $p_i, i \in [1,N_{problem}]$ and  a skill embedding $s_i, i \in [1,N_{skill}]$ to vectorize problems and skills. $N_{problem}$ and $N_{skill}$ are the number of problems and skills, respectively.
Similarly, the pairs $\{(\pi_i,r_i)\}_{i=1}^{N_{seq}}$ are also converted to embedding $k_j, j \in [1,2*N_{skill}]$, where $j=\pi_i + r_i * N_{problem}$. The $k_j$ denotes the updates on user knowledge after observing student interaction $(\pi_i,r_i)$.

In the proposed model, we first obtained problem embedding sequences $\{p_{\pi i}\}_{i=1}^{N_{seq}}$ and skill embedding sequences $\{s_{\pi i}\}_{i=1}^{N_{seq}}$ corresponding to input problem solving history $\{\pi_i\}_{i=1}^{N_{seq}}$. With these, the model aims to produce the probability of student making correct answer, $\hat{r_i}$ which is the model prediction corresponding to $r_i$.

\subsection{Skill2Vec Input for Knowledge Tracing}
As mentioned in the dataset description, skill-to-skill relationships are given as a graph $G=(V,E)$; the node $V = \{v_i\}, i \in [1,N_{skill}]$ are the skills, and the edge $E = \{v_i,v_j\}, i,j \in [1,N_{skill}]$ denote whether the skills have relationships.
Because the relationships between skills are represented as a graph, directly using skill-to-skill relationships as a knowledge tracing input is impossible. 
Therefore, we first need to convert the graph into the proper format for input into the knowledge tracing model.
Figure \ref{fig:overview}-(a) illustrates the conversion.
As the figure shows, we chose to convert the graph into vectors of skills $\{s_i\}, i \in [1,N_{skill}]$. To vectorize the graph, we utilized Node2Vec\cite{node2vec}, which learns the best vector representation of the nodes by maximizing the likelihood of neighborhood nodes.
After converting the graph, we constructed a series of vectors using skill vectors that corresponds to each problem in problem-solving sequences.
We refer to the resulting vector sequence, $\{s_{\pi_{i}}\}_{i=1}^{N_{seq}}$, as Skill2Vec.

\subsection{Skill Projection Loss}
Properly using Skill2Vec for knowledge tracing is as important as making it.
A typical way of using the feature is simply adding the Skill2Vec feature to the input problem embedding.
However, this approach has some ambiguity in terms of the addition factors and may damage crucial information for knowledge tracing as it directly modifies problem features.
Therefore, we propose a skill projection loss ($L_p$) which encourages the problem features to contain skill-to-skill relation knowledge while maintaining other information for knowledge tracing.
Figure \ref{fig:overview}-(b) illustrates the $L_p$ loss. When the sequence of problem embedding $\{p_{\pi_{i}}\}_{i=1}^{N_{seq}}$ is given, we produce a projected problem embedding sequence $\{p_{\pi_{i}}'\}_{i=1}^{N_{seq}}$ by passing the sequence to the projection layer. Then, we calculate the mean squared error of $p_i'$ and $s_i$ where $s_i$ is the Skill2Vec feature.

The skill projection loss can be formulated as follows.
\begin{equation}
    L_p = \frac{1}{N_p} \sum_{i=1}^{N_p} (p_i' - s_{pi})
    \label{equation:projectionloss}
\end{equation}

\subsection{Deep Knowledge Tracing Model}
We use two embedding sequences for knowledge tracing. The first is designed to embed knowledge about problems ($\{p_{\pi_{i}}\}_{i=1}^{N_{seq}}$), and the other embeds the updates on student knowledge state ($\{k_{\pi_{i}}\}_{i=1}^{N_{seq}}$).
For knowledge tracing, we simply employ transformer architecture \cite{transformer}. The input source embedding is $\{p_{\pi_{i}}\}_{i=1}^{N_{seq}}$, and the input target embedding is $\{k_{\pi_{i}}\}_{i=1}^{N_{seq}}$, similar to SAINT \cite{saint}.
The output of the transformer is then passed to simple linear layer, to produce model prediction $\hat{r_i}$.

With model prediction $\hat{r_i}$ and actual student correctness $r_i$, the knowledge tracing loss $L_k$ is calculated as follows:
\begin{equation}
    L_k = - \frac{1}{N_{seq}} \sum_{i=1}^{N_{seq}} {(r_i\log(\hat{r_i}) + (1 - r_i)\log(1 - \hat{r_i}))}
    \label{equation:ktloss}
\end{equation}

The final loss function of our method is as follows:

\begin{equation}
    L = L_k + \lambda L_p
    \label{equation:loss}
\end{equation}

 Above, $\lambda$ is a balancing constant between $L_k$ and $L_p$.

\section{Experiment}

\begin{figure}[t!] 
    \centerline{\includegraphics[width=0.45\textwidth]{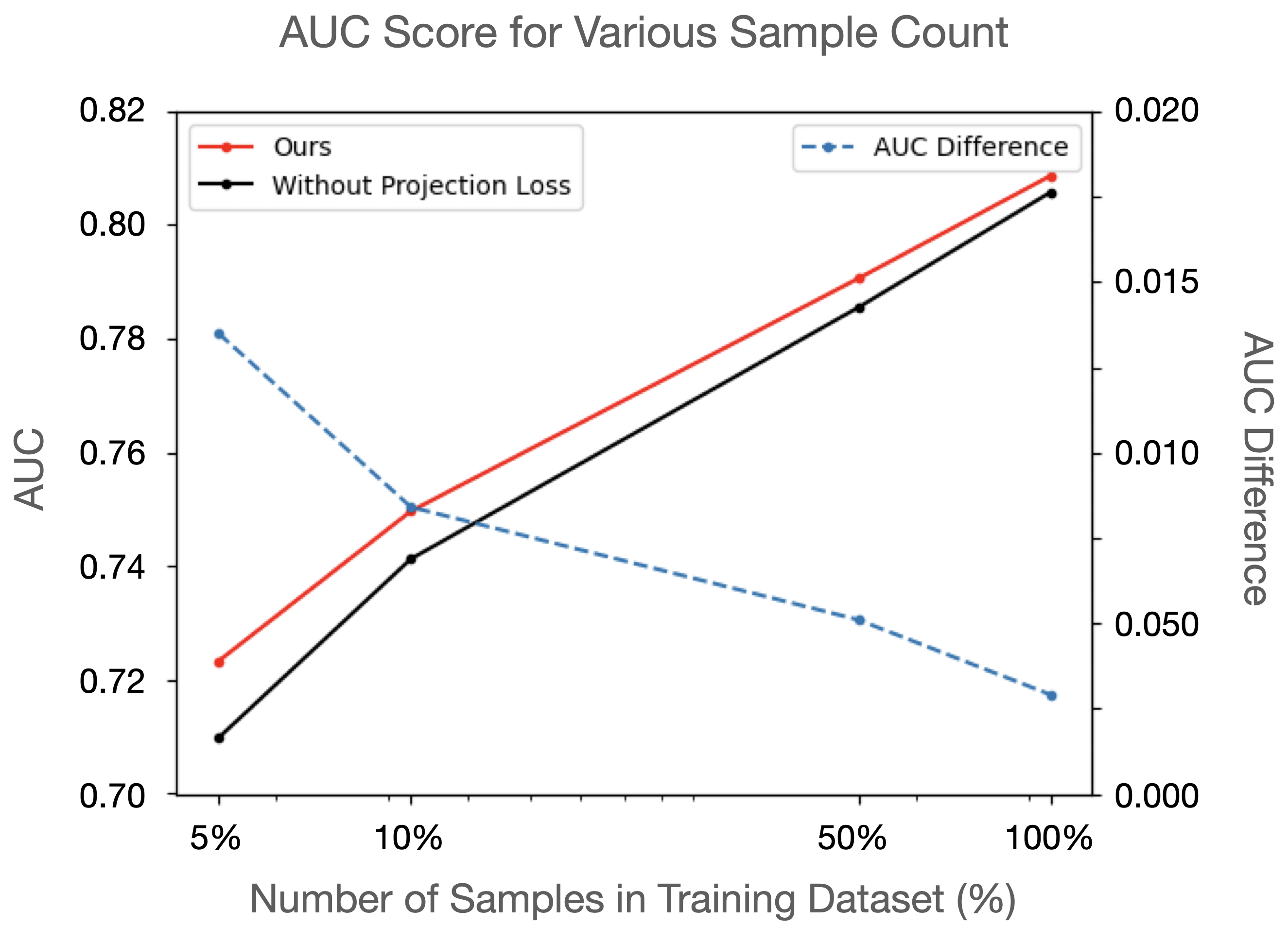}}
    \caption{Change in AUC scores with training dataset size.}
	\label{fig:small}
\end{figure}

\subsection{Experimental Details}

For Skill2vec, we used a vector with a 25-dimensional skill embedding. While making that embedding using the Node2Vec method, we fixed the walk length to 128, p, and q values to 1. Additionally, we set the number of walks value as 300,000 and the window value as 4.

For the model, we employed a simple, fully connected layer for projection and set the hidden layer size to 100.
For the Transformer layer, we used one head, four encoder layers, and four decoder layers.
For every fully connected layer, we applied dropout with p=0.05.
We used Adam optimizer with a learning rate of 0.0002 when training the knowledge tracing model.

\subsection{Results}

\begin{table}[htb]
\caption{Results}
\centering
\begin{tabular}{cc}
\toprule
Method & AUC \\
\midrule
 Without Projection Loss   & 80.81  \\
 Random Skill to Skill  & 80.90 \\
 Ours    & \textbf{81.10}  \\
\bottomrule
\end{tabular}
\label{table:result}
\end{table}

Table \ref{table:result} is the model performance on ASSIST2009-SSR dataset. 
In the table, "Without Projection Loss" corresponds to the proposed model without projection loss, and "Random Knowledge" refers to the model with random skill-to-skill relationships. 
The proposed model outperformed the model without projection loss. This suggests that the expert labeled skill to skill relationships can help improve the knowledge tracing performance.
Also, the proposed model achieved higher performance compared to the model with a random skill-to-skill graph. 
This indicates that the quality of labeled skill-to-skill relationships is important. In other words, knowledge from a skilled expert can improve a knowledge tracing system.

\subsection{Results on Limited Dataset}
Because expert knowledge is not affected by the size of a given dataset, it could be more advantageous when the dataset is small.
To demonstrate this, we trained the proposed model and the model without projection loss on 5\%, 10\%, 50\% of the original training dataset.
Figure \ref{fig:small} shows the performance with the size of the dataset. As the dataset becomes small, the AUC margin between the proposed model and the model without loss increased.

\section{Conclusion}
To the best of our knowledge, the present work is the first to incorporate expert labeled skill-to-skill relationships in knowledge tracing.
We constructed skill-to-skill relationship data and proposed a novel method that can utilize the data.
Empirical results show that the proposed methods improve AUC on a real-world dataset. This indicates that incorporating human domain knowledge into knowledge tracing methods can play a key role in constructing better knowledge tracing models.
Moreover, our results on limited datasets show that the gap between the model proposed in this paper and the baseline model (the model without projection loss) increased as the number of samples that the model used decreased. 
From this result, we can ascertain that expert knowledge between skills guides the model in the right direction.
This means that our data on skill-to-skill relationships can provide insights that can be gained from a large dataset, and our model leveraged this information successfully.

\section{Acknowledgement}
We thank Insu Lee, Hongsik Tae, Jiin Kim, and Misook Kim for their invaluable contributions to the research through their annotation efforts. This work was supported by the National Research Foundation of Korea(NRF) grant funded by the Korea government(MSIT) (No. 2022R1A4A3033874), and supported by the National Research Foundation of Korea(NRF) grant funded by the Korea government(MSIT). (No.2021R1F1A1060117).

\bibliography{aaai23} 

\bigskip

\end{document}